\documentclass[runningheads]{llncs}

\usepackage{eccv}

\usepackage{eccvabbrv}

\usepackage{graphicx}
\usepackage{booktabs}

\usepackage[accsupp]{axessibility}  

\usepackage{hyperref}

\usepackage{orcidlink}

\begin{document}

\title{One-Shot Learning Meets Depth Diffusion in Multi-Object Videos}


\author{Anisha Jain}

\authorrunning{}

\institute{Carnegie Mellon University}

\maketitle

\begin{abstract}

Creating editable videos that depict complex interactions between multiple objects in various artistic styles has long been a challenging task in filmmaking. Progress is often hampered by the scarcity of data sets that contain paired text descriptions and corresponding videos that showcase these interactions. This paper introduces a novel depth-conditioning approach that significantly advances this field by enabling the generation of coherent and diverse videos from just a single text-video pair using a pre-trained depth-aware Text-to-Image (T2I) model. Our method fine-tunes the pre-trained model to capture continuous motion by employing custom-designed spatial and temporal attention mechanisms. During inference, we use the DDIM inversion to provide structural guidance for video generation. This innovative technique allows for continuously controllable depth in videos, facilitating the generation of multiobject interactions while maintaining the concept generation and compositional strengths of the original T2I model across various artistic styles, such as photorealism, animation, and impressionism.

\end{abstract}

\section{Introduction}
\label{sec:intro}

Recent advances in text-to-image (T2I) models have showcased remarkable prowess in crafting lifelike images from textual prompts \cite{rombach2022highresolution}. In order to replicate this achievement in Text-to-Video (T2V) synthesis, recent efforts \cite{ho2022video,ho2022imagen,zhou2023magicvideo,singer2022makeavideo} have expanded spatial-only T2I models into the spatio-temporal realm. Typically, these models adhere to the conventional approach of training in large text-video corpora (e.g., WebVid-10M \cite{bain2022frozen}). While producing promising outcomes in T2V synthesis, this methodology demands substantial training resources on robust hardware accelerators, incurring significant costs and time investments.

Our daily interactions frequently involve multiple objects, and our engagement is often dictated by their relative spatial arrangements. For example, when driving, awareness of the positions of surrounding vehicles is vital. This task proves challenging due to the dynamic and interactive nature of objects in motion. To navigate such interactions effectively, we rely on estimating distances from objects and adjusting our actions accordingly, underscoring the pivotal role of depth perception.

Given the ubiquitous nature of interactions with multiple objects, it becomes imperative to extend the capabilities of T2V models to generate videos featuring such interactions. This endeavor presents challenges due to the heightened temporal intricacies. Generating videos depicting multiple objects' interactions demands the model's comprehension of both spatial and temporal dynamics, ensuring coherent video synthesis amidst occlusions and increased complexity.

In this study, our objective is to produce high-quality videos that illustrate object interactions. We propose an innovative one-shot training approach capable of generating cohesive videos depicting multiple objects engaging with each other. Our methodology is based on the foundation of a pre-trained depth-conditioned T2I model, as outlined in \cite{rombach2022highresolution}. However, employing full spatiotemporal attention invariably results in a quadratic increase in computational complexity, rendering it impractical for generating videos with expanding frame counts. Furthermore, adopting a simplistic fine-tuning strategy that updates all parameters risks compromising the existing knowledge encoded within T2I models, thus impeding the generation of videos depicting novel concepts. To address these challenges, we introduce a sparse spatio-temporal attention mechanism, limiting visits to only the initial and preceding video frames. Additionally, we implement an efficient tuning strategy focused solely on updating the projection matrices within the attention blocks. Empirical evaluations demonstrate that these innovations maintain object consistency across frames, albeit with limited continuous motion. To overcome this limitation during inference, we use structure guidance from input videos through DDIM inversion, a process described in \cite{song2022denoising}. By initializing noise with the inverted latent information, we achieve temporally coherent videos characterized by fluid motion.

In summary, our proposed approach excels in producing high-quality videos portraying object interactions by utilizing a one-shot training method that relies on just one text-video pair. Additionally, our model inherits the versatile composition skills from pre-trained T2I models, guaranteeing the creation of fresh ideas. Consequently, it can generate a variety of objects, backgrounds, and interactions while preserving consistency across frames. Moreover, our technique ensures smooth motion in the generated videos, resulting in the creation of temporally cohesive content.

\section{Related Work}
\label{sec:related}
Our focus revolves around the convergence of various disciplines: diffusion models and techniques for creating images/videos based on textual cues, text-guided manipulation of actual visual content, and generative models refined using a singular video data set. In this summary, we outline the notable achievements within each domain, emphasizing their correlations and distinctions compared to our proposed approach.

\subsection{Text-to-Image Synthesis}
\label{sec:t2i}
Research in Text-to-Image (T2I) generation has undergone extensive exploration, and many earlier models rely primarily on transformer architectures \cite{pmlr-v139-ramesh21a,yu2022scaling,ding2022cogview2,gafni2022makeascene,yu2022vectorquantized}. Recently, several T2I generative models \cite{nichol2022glide, saharia2022photorealistic, gu2022vector,rombach2022highresolution} have changed to using diffusion models \cite{ho2020denoising}. For example, GLIDE \cite{nichol2022glide} introduces classifier-free guidance within the diffusion model \cite{ho2022classifierfree} to improve image fidelity, while DALLE-2 \cite{ramesh2022hierarchical} improves text-image alignment by using CLIP's feature space \cite{radford2021learning}. Imagen \cite{ho2022imagen} employs cascaded diffusion models for high-definition video generation, and subsequent advances such as VQdiffusion \cite{gu2022vector} and Latent Diffusion Models (LDMs) \cite{rombach2022highresolution} operate within the latent space of an auto-encoder to enhance training efficiency. LDMs have shown effectiveness in fine-tuning for conditioned outputs. Our approach extends upon depth-conditioned LDMs by extending the 2D model into the spatio-temporal domain within the latent space.

\subsection{Text-to-Video Synthesis}
\label{sec:t2v}

Creating realistic videos poses a significant challenge due to their intricate and high-dimensional structures. Initially, attention was paid to Generative Adversarial Networks (GANs) \cite{goodfellow2014generative}, which produce samples from a Gaussian distribution through a two-player game. However, GAN-based methods face training difficulties, especially with large datasets. Recent advances, leveraging large language models \cite{radford2021learning} and transformers \cite{vaswani2023attention}, focus on generating videos from text descriptions. For example, Wu et al. \cite{wu2021godiva} extend VQ-VAE \cite{oord2018neural} for text-to-video synthesis by mapping text tokens to video tokens. N\"{U}WA \cite{wu2021nuwa} introduces an auto-regressive framework applicable to both text-to-image and video generation. Enhancements in video quality are achieved through approaches like CogVideo \cite{hong2022cogvideo}, which integrates temporal attention modules and pretrained text-to-image models for Text-to-Video (T2V) synthesis. The diffusion-based method gains traction, with Video Diffusion Models (VDM) \cite{ho2022video} using a factorized space-time U-Net for direct pixel diffusion. Further refinements are seen in Imagen Video \cite{ho2022imagen}, which improves on VDM with cascaded diffusion models and parameterization for prediction. Similar progress is made by Make-A-Video \cite{singer2022makeavideo} in extending diffusion-based models from text-to-image generation \cite{rombach2022highresolution} to T2V synthesis. However, these models require extensive training on large video datasets, which limits their scalability and practicality. Another work, Tune-A-Video \cite{wu2023tuneavideo}, extends image synthesis diffusion models to multi-object video generation by introducing depth conditioning and temporal connections into a pre-existing image model. But these models fail in generating videos with multiple objects and occlusions. 

\subsection{Depth-guided Image and Video Synthesis}
\label{sec:depth}
Employing a morphological diffusion-based technique \cite{Guo2008ASI}, \cite{7471866} presents an approach to seamlessly complete depth maps, ensuring smooth structural propagation across undesired regions. These inferred depth values then serve as a guide for filling in missing texture in the corresponding color image. Latent Diffusion Models (LDMs) \cite{rombach2022highresolution} can be adapted to incorporate depth maps as conditioning variables, enabling the generation of high-resolution images.

DVI \cite{liao2020dvi} introduces a novel depth-guided video inpainting model that leverages depth maps to steer the inpainting process, synthesizing missing content by amalgamating information from multiple videos. Meanwhile, DCVGAN \cite{8803764} is a depth-conditioned video generation model employing a conditional GAN architecture to produce videos from depth maps, trained on a dataset comprising paired depth maps and real-world videos. However, generative models face challenges during training. GD-VDM \cite{lapid2023gdvdm} adopts a two-phase generation strategy that involves the generation of depth videos followed by a new Vid2Vid diffusion model to produce coherent real-world videos, although it requires extensive training on large datasets.

In line with the approach of Tune-A-Video \cite{wu2023tuneavideo}, we extend image synthesis diffusion models to facilitate controllable multiobject video generation by incorporating depth conditioning and temporal connections into an existing image model.

\begin{figure}[!ht]
    \centering
    \resizebox{\textwidth}{!}{\includegraphics[width=\textwidth]{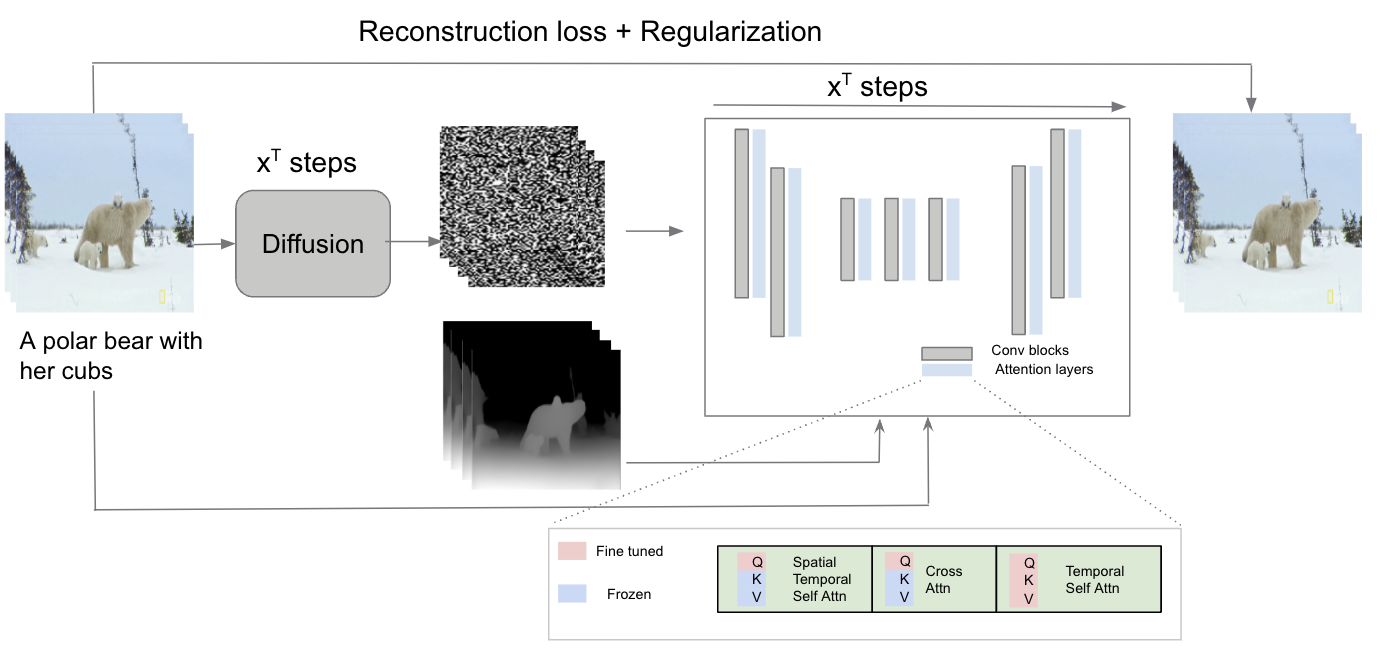}}
    \caption{Using a text-video pair (for example, “a polar bear with her cubs”) as input, our approach utilizes pretrained depth-conditioned T2I diffusion models to generate T2V content. During fine-tuning, we update the projection matrices in attention blocks using the standard diffusion training loss and regularization.}
    \label{fig:training-pipeline}
\end{figure}

\section{Methodology}
\label{sec:method}

\begin{figure}[ht!]
    \centering
    \resizebox{\textwidth}{!}{
    \includegraphics[width=\textwidth]{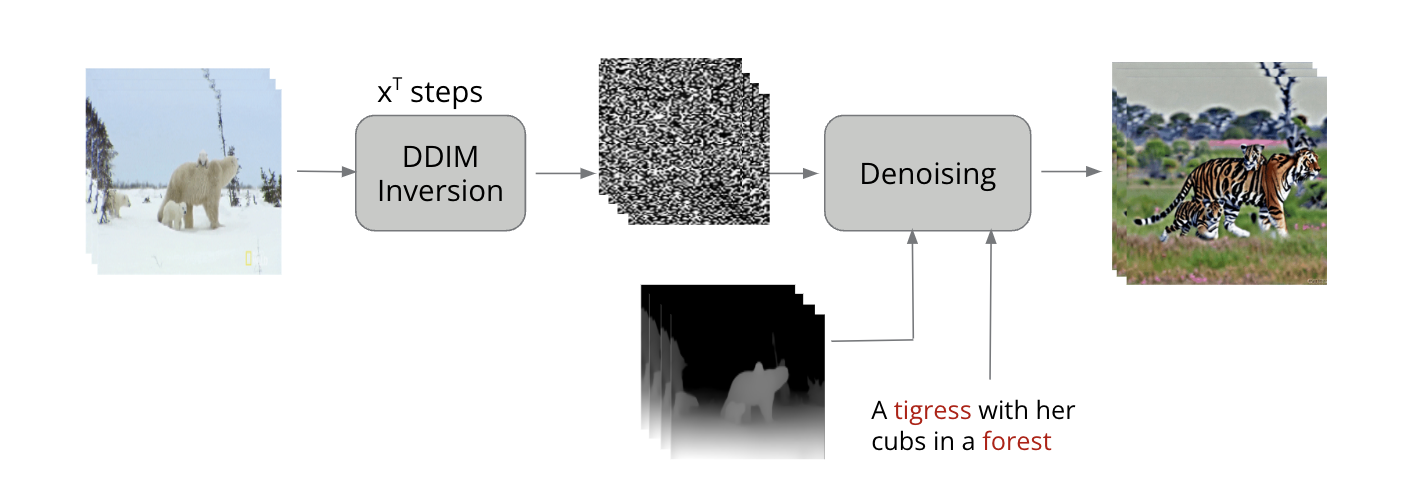}}
    \caption{During inference, we generate a new video by sampling from the latent noise inverted from the input video, using an edited prompt (e.g., “A tigress with her cubs in a forest”) as guidance.}
    \label{fig:inference-pipeline}
\end{figure}

Our objective in the text-guided depth controllable video generation task is to produce lifelike videos using both a video sequence and textual descriptions of appearance as input. Our approach is based on the foundation of a pretrained depth-conditioned text-to-image model, specifically the latent diffusion model \cite{rombach2022highresolution}, with several customized adjustments to align with our task requirements. \ref*{sec:prelim} provides a brief overview of the preliminary concepts, followed by a detailed explanation of our depth-guided text-to-video generation approach in \ref{sec: depth-guided}.

\subsection{Premilinary: Latent Diffusion Model}
\label{sec:prelim}
Latent diffusion models (LDM) \cite{rombach2022highresolution} represent a category of diffusion models that characterize the latent space distribution of images, showcasing notable advancements in image synthesis. This model comprises two main components: an autoencoder and a diffusion model. The autoencoder, composed of an encoder $\mathcal{E}$ and a decoder $\mathcal{D}$, is trained to reconstruct images. Specifically, the encoder projects the input images $x$ into a lower-dimensional latent space: $z = \mathcal{E}(x)$, while the decoder reconstructs the original image from the latent space: $\tilde{x} = \mathcal{D}(z)$. On the other hand, the diffusion model learns the distribution of the latent space of images $z_0 \sim p_{\text{data}}(z_0)$ using the diffusion denoising probabilistic model (DDPM) \cite{ho2020denoising} and generates new samples within the latent space.

The generation process involves a gradual backward denoising process over $T$ time steps, beginning from pure Gaussian noise $z_T$ and ending at a novel sample $z_0$. Mathematically, this process is defined as follows:
\begin{equation}
T p_{\theta}(z_{0:T}) := p(z_T) \prod_{t=1}^{T} p_{\theta}(z_{t-1}|z_t),
\end{equation}
where
\begin{equation}
p_{\theta}(z_{t-1}|z_t) := \mathcal{N}(z_{t-1}; \mu_{\theta}(z_t, t), \Sigma_{\theta}(z_t, t)).
\end{equation}
Conversely, the Markov chain progresses through a gradual forward noising process using a predefined noise schedule $\beta_1, \ldots, \beta_T$, expressed as:
\begin{equation}
T q(z_{1:T}|z_0) := q(z_1|z_0) \prod_{t=1}^{T} q(z_{t+1}|z_t),
\end{equation}
where
\begin{equation}
q(z_{t+1}|z_t) := \mathcal{N}(z_{t+1}; (1 - \beta_t)z_t, \beta_t I).
\end{equation}
During each timestep, random noise $\epsilon$ is sampled from a diagonal Gaussian distribution, and a time-conditioned denoising model $\theta$ is trained to predict the added noise at each timestep using mean squared error (MSE) loss:
\begin{equation}
L(\theta) := |\epsilon - \epsilon_{\theta}(z_t, t)|^2_2.
\end{equation}

\subsection{Depth-guided Text-to-Video Generation}
\label{sec:depth-guided}

We utilize depth-conditioned LDM as the backbone for our text-guided depth controllable video generation task. Following a similar approach as \cite{wu2023tuneavideo}, we expand the base model. This entails extending the 2D LDM to the spatio-temporal realm by transforming the 2D convolution layers into pseudo-3D convolution layers. Here, $3 \times 3$ kernels are replaced by $1 \times 3 \times 3$ kernels, and we introduce a temporal self-attention layer within each transformer block for temporal modeling. To bolster temporal coherence, we introduce a sparse spatiotemporal attention mechanism that selectively focuses on initial and preceding video frames. During fine-tuning, we adopt a strategy that concentrates solely on updating the projection matrices within the attention blocks. This ensures retention of knowledge from the pretrained T2I model, facilitating the generation of innovative concepts.

The spatiotemporal attention mechanism is devised to uphold temporal consistency by referencing pertinent positions in previous frames. Consequently, the key and value matrices remain fixed, with only the query matrices being updated during fine-tuning. Mathematically, the attention mechanism is expressed as follows.

\begin{equation}
\text{Attention}(Q, K, V) = \text{softmax}\left(\frac{QK^T}{\sqrt{d_k}}\right)V,
\end{equation}

where $Q$, $K$, and $V$ denote the query, key, and value matrices, respectively, and $d_k$ represents the dimensionality of the key vectors. These matrices are defined as follows.

\begin{equation}
Q = W^Q z_{v_i}, \quad K = W^K [z_{v_1} z_{v_{i-1}}], \quad V = W^V [z_{v_1} z_{v_{i-1}}],
\end{equation}

where $W^Q$, $W^K$, and $W^V$ are the projection matrices, and $z_{v_i}$ denotes the latent representation of the current video frame, while $z_{v_1}$ and $z_{v_{i-1}}$ represent the latent representations of the first and previous video frames, respectively. In particular, the projection matrices are shared across spatial and temporal dimensions. Additionally, the newly introduced temporal self-attention layers are fully fine-tuned during training. To refine text-video alignment, the query projection matrix in cross-attention layers is updated during fine-tuning. This strategy of fine-tuning only the projection matrices within the attention blocks ensures retention of knowledge encoded within the pretrained T2I model, facilitating the generation of novel concepts.

\textbf{Loss and Regularization:} We employ the same training objective as LDM, augmented with a temporal consistency loss to ensure smooth motion in the generated videos. The loss term is defined as:

\begin{equation}
L_{\text{temporal}} = \sum_{i=1}^{T-1} \left| z_{v_i} - z_{v_{i+1}} \right|_2^2,
\end{equation}

where $T$ represents the total number of video frames. This loss encourages the latent representations of consecutive video frames to be similar, thereby promoting smooth motion in the generated videos.

\textbf{Inference:} During the inference phase, we employ DDIM inversion to provide structural guidance. We incorporate structural cues from the source video during inference. Specifically, we obtain a latent noise of the source video $\mathcal{V}$ through DDIM inversion without textual conditions. This noise serves as the initial point for DDIM sampling, guided by an edited prompt $\mathcal{T}^{\ast}$. The resulting video $\mathcal{V}^{\ast}$ is then obtained as $\mathcal{V}^{\ast} = \mathcal{D}(\text{DDIM-samp}(\text{DDIM-inv}(\mathcal{V}), \mathcal{T}^{\ast}))$.

\begin{figure}[!ht]
    \resizebox{\textwidth}{!}{\includegraphics{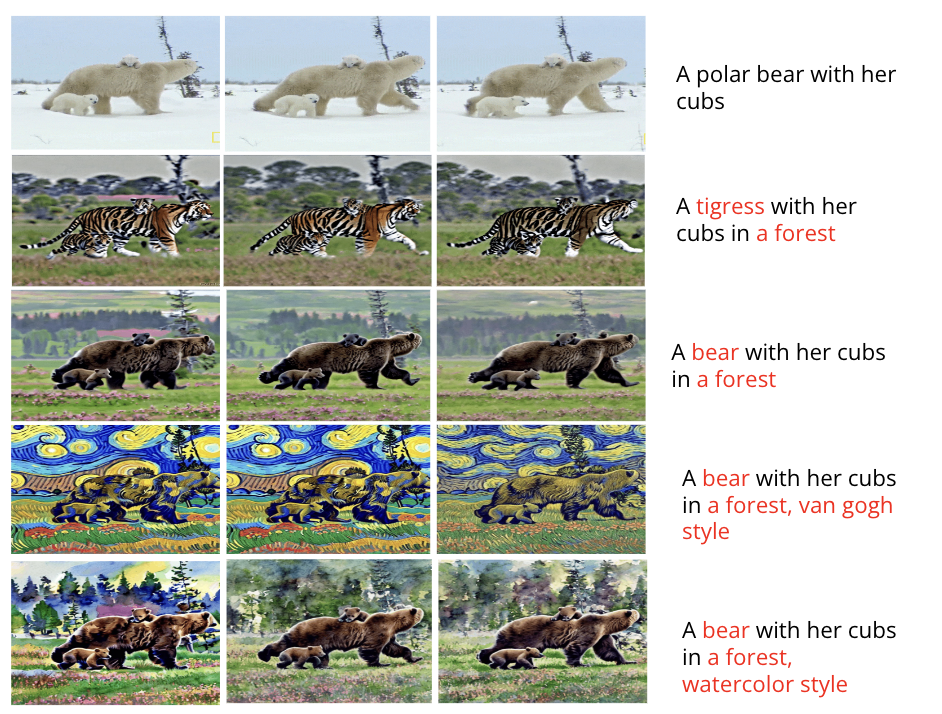}}
    \caption{Sample results of our method. The first row consists of images from the input sequence along with the text prompt. The rows below consist of images from generated sequences for the edited prompts for different styles and novelty.}
    \label{fig:our-result}
\end{figure}

\section{Experiments}

\subsection{Implementation Details}
Our work utilizes Latent Diffusion Models \cite{rombach2022highresolution} along with publicly available pre-trained weights \footnote{\url{https://huggingface.co/stabilityai/stable-diffusion-2-depth}}. We extract 8-12 evenly spaced frames from the input video at a resolution of 512 × 512, and then fine-tune the models using our approach for 500 steps with a learning rate of $1 \times 10^{-5}$ and a batch size of 1. During inference, we employ the DDIM sampler \cite{song2022denoising} and classifier-free guidance \cite{ho2022classifierfree} in our tests. Fine-tuning a single video takes approximately 10 minutes, while sampling requires about 1 minute on an NVIDIA A100 GPU.

\subsection{Qualitative Results}
\begin{figure}[ht!]
    \includegraphics[width=0.5\textwidth]{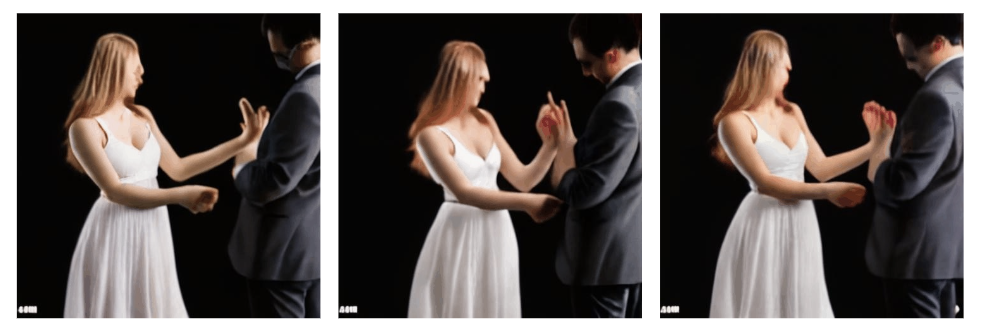}
    \includegraphics[width=0.5\textwidth]{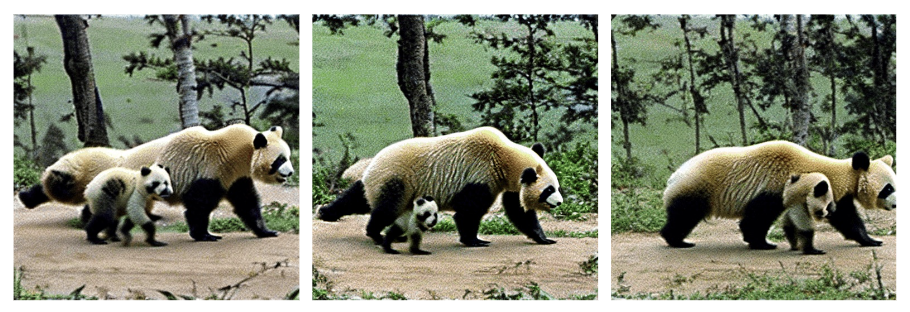}
    \caption{On the left, Cog-Video produces undesirable results for human interactions with blurred faces and weird hands. While on the right, Tune-A-Video produces unpleasant results when the input video contains multiple objects and exhibits occlusions. For example, the two pandas at the bottom being mixed together. }
    \label{fig:their-video-result}
\end{figure}

We present a visual comparison of our proposed approach against two baselines, Tune-a-Video and CogVideo. The results of this work can be found in \ref{fig:their-video-result}. The differences can be observed starkly with the results of our work, as in \ref{fig:our-result}.

\subsection{Quantitative Comparison}
We evaluated our method against the baselines using automatic metrics and a user study, focusing on the consistency of the frames and the textual faithfulness, as detailed in \ref{tab:comparison}.

For automatic metrics, frame consistency was measured by computing CLIP\cite{radford2021learning} image embeddings for all video frames and reporting the average cosine similarity between frame pairs. Textual faithfulness was assessed by calculating the average CLIP score between video frames and their corresponding edited prompts. Our method outperformed the baselines, with CogVideo showing consistent frames but poor textual representation, and Tune-a-Video achieving high textual faithfulness but inconsistent content.

\begin{table}[h]
\centering
\begin{tabular}{lcccc}
\toprule
Method & \multicolumn{2}{c}{Frame Consistency} & \multicolumn{2}{c}{Textual Alignment} \\
\cmidrule(r){2-3} \cmidrule(r){4-5}
 & CLIP Score & User Preference & CLIP Score & User Preference \\
\midrule
CogVideo & 90.64 & 12.14 & 23.91 & 15.00 \\
Tune-A-Video & 92.40 & 45.64 & 27.58 & 35.52 \\
Ours & \textbf{94.73} & \textbf{90.63*} / \textbf{74.87**} & \textbf{29.41} & \textbf{87.30*} / \textbf{78.08**} \\
\bottomrule
\end{tabular}
\caption{Comparison of Methods for Frame Consistency and Textual Alignment. * indicates Ours vs Cog-Video, ** indicates Ours vs Tune-a-Video}
\label{tab:comparison}
\end{table}

\begin{figure}[!ht]
    \centering
    \includegraphics[width=\textwidth]{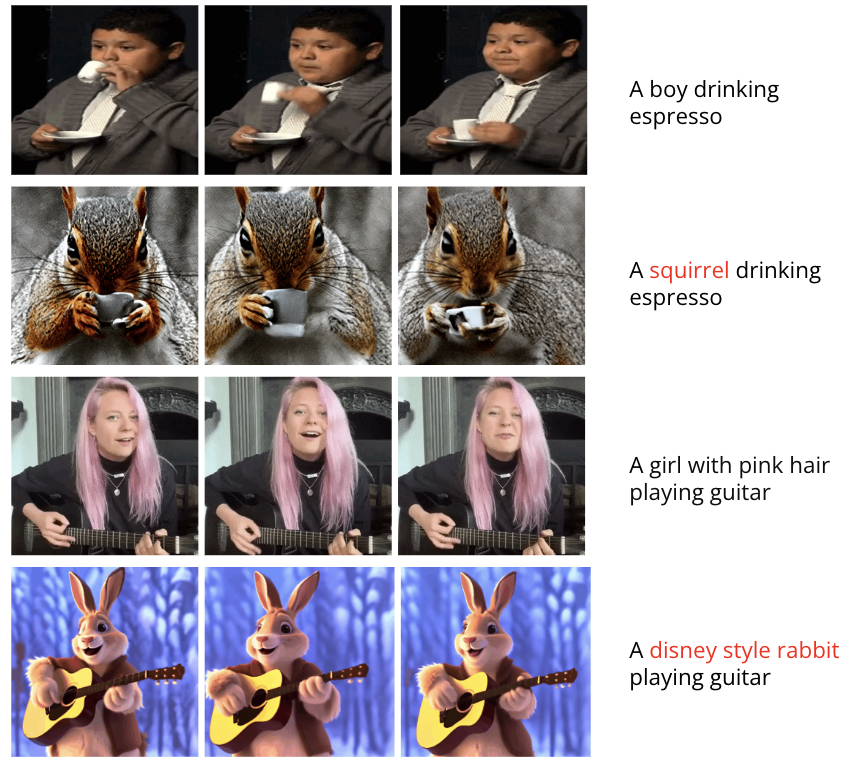}
    \caption{Some more results of our method. The first and third row show frames from the input sequence. The second and fourth row illustrate the frames from the respective generated sequences for the edited prompt.}
    \label{fig:our-result}
\end{figure}

In the user study, participants were shown two videos (one from our method and one baseline) in random order and asked to select the one with better temporal consistency. For textual faithfulness, participants were shown the textual description and asked which video aligned better with it. Five participants rated each example, with the final result determined by majority vote. Our method was preferred over CogVideo and Tune-a-Video for both frame consistency and textual faithfulness.

\section{Conclusion}
\label{sec:conclusion}
This work demonstrates that depth conditioning significantly enhances text-guided video generation using Latent Diffusion Models (LDM). Although previous research used LDMs for video generation, we found that depth conditioning improves the quality of videos, particularly in scenes with multiple interacting objects. By extending 2D LDMs to the spatio-temporal domain and incorporating pseudo 3D convolutions and temporal self-attention, we achieved better temporal coherence.

Our fine-tuning strategy, updating only projection matrices, retains pretrained T2I model knowledge while generating novel concepts. The loss of temporal consistency ensures smooth motion and the DDIM inversion enhances video quality. Depth conditioning proves crucial for effective multi-object interaction modeling, representing a significant advancement in video generation. Future work will explore generation of longer videos with more complex scenes and improve model efficiency.


%
%
\bibliographystyle{splncs04}
\bibliography{main}

\begin{thebibliography}{10}
\providecommand{\url}[1]{\texttt{#1}}
\providecommand{\urlprefix}{URL }
\providecommand{\doi}[1]{https://doi.org/#1}

\bibitem{bain2022frozen}
Bain, M., Nagrani, A., Varol, G., Zisserman, A.: Frozen in time: A joint video and image encoder for end-to-end retrieval (2022)

\bibitem{7471866}
Ciotta, M., Androutsos, D.: Depth guided image completion for structure and texture synthesis. In: 2016 IEEE International Conference on Acoustics, Speech and Signal Processing (ICASSP). pp. 1199--1203 (2016). \doi{10.1109/ICASSP.2016.7471866}

\bibitem{ding2022cogview2}
Ding, M., Zheng, W., Hong, W., Tang, J.: Cogview2: Faster and better text-to-image generation via hierarchical transformers (2022)

\bibitem{gafni2022makeascene}
Gafni, O., Polyak, A., Ashual, O., Sheynin, S., Parikh, D., Taigman, Y.: Make-a-scene: Scene-based text-to-image generation with human priors (2022)

\bibitem{goodfellow2014generative}
Goodfellow, I.J., Pouget-Abadie, J., Mirza, M., Xu, B., Warde-Farley, D., Ozair, S., Courville, A., Bengio, Y.: Generative adversarial networks (2014)

\bibitem{gu2022vector}
Gu, S., Chen, D., Bao, J., Wen, F., Zhang, B., Chen, D., Yuan, L., Guo, B.: Vector quantized diffusion model for text-to-image synthesis (2022)

\bibitem{Guo2008ASI}
Guo, H., Ono, N., Sagayama, S.: A structure-synthesis image inpainting algorithm based on morphological erosion operation. 2008 Congress on Image and Signal Processing  \textbf{3},  530--535 (2008), \url{https://api.semanticscholar.org/CorpusID:17003954}

\bibitem{ho2022imagen}
Ho, J., Chan, W., Saharia, C., Whang, J., Gao, R., Gritsenko, A., Kingma, D.P., Poole, B., Norouzi, M., Fleet, D.J., Salimans, T.: Imagen video: High definition video generation with diffusion models (2022)

\bibitem{ho2020denoising}
Ho, J., Jain, A., Abbeel, P.: Denoising diffusion probabilistic models. arXiv preprint arxiv:2006.11239  (2020)

\bibitem{ho2022classifierfree}
Ho, J., Salimans, T.: Classifier-free diffusion guidance (2022)

\bibitem{ho2022video}
Ho, J., Salimans, T., Gritsenko, A., Chan, W., Norouzi, M., Fleet, D.J.: Video diffusion models (2022)

\bibitem{hong2022cogvideo}
Hong, W., Ding, M., Zheng, W., Liu, X., Tang, J.: Cogvideo: Large-scale pretraining for text-to-video generation via transformers (2022)

\bibitem{lapid2023gdvdm}
Lapid, A., Achituve, I., Bracha, L., Fetaya, E.: Gd-vdm: Generated depth for better diffusion-based video generation (2023)

\bibitem{liao2020dvi}
Liao, M., Lu, F., Zhou, D., Zhang, S., Li, W., Yang, R.: Dvi: Depth guided video inpainting for autonomous driving (2020)

\bibitem{8803764}
Nakahira, Y., Kawamoto, K.: Dcvgan: Depth conditional video generation. In: 2019 IEEE International Conference on Image Processing (ICIP). pp. 749--753 (2019). \doi{10.1109/ICIP.2019.8803764}

\bibitem{nichol2022glide}
Nichol, A., Dhariwal, P., Ramesh, A., Shyam, P., Mishkin, P., McGrew, B., Sutskever, I., Chen, M.: Glide: Towards photorealistic image generation and editing with text-guided diffusion models (2022)

\bibitem{oord2018neural}
van~den Oord, A., Vinyals, O., Kavukcuoglu, K.: Neural discrete representation learning (2018)

\bibitem{radford2021learning}
Radford, A., Kim, J.W., Hallacy, C., Ramesh, A., Goh, G., Agarwal, S., Sastry, G., Askell, A., Mishkin, P., Clark, J., Krueger, G., Sutskever, I.: Learning transferable visual models from natural language supervision (2021)

\bibitem{ramesh2022hierarchical}
Ramesh, A., Dhariwal, P., Nichol, A., Chu, C., Chen, M.: Hierarchical text-conditional image generation with clip latents (2022)

\bibitem{pmlr-v139-ramesh21a}
Ramesh, A., Pavlov, M., Goh, G., Gray, S., Voss, C., Radford, A., Chen, M., Sutskever, I.: Zero-shot text-to-image generation. In: Meila, M., Zhang, T. (eds.) Proceedings of the 38th International Conference on Machine Learning. Proceedings of Machine Learning Research, vol.~139, pp. 8821--8831. PMLR (18--24 Jul 2021)

\bibitem{rombach2022highresolution}
Rombach, R., Blattmann, A., Lorenz, D., Esser, P., Ommer, B.: High-resolution image synthesis with latent diffusion models (2022)

\bibitem{saharia2022photorealistic}
Saharia, C., Chan, W., Saxena, S., Li, L., Whang, J., Denton, E., Ghasemipour, S.K.S., Ayan, B.K., Mahdavi, S.S., Lopes, R.G., Salimans, T., Ho, J., Fleet, D.J., Norouzi, M.: Photorealistic text-to-image diffusion models with deep language understanding (2022)

\bibitem{singer2022makeavideo}
Singer, U., Polyak, A., Hayes, T., Yin, X., An, J., Zhang, S., Hu, Q., Yang, H., Ashual, O., Gafni, O., Parikh, D., Gupta, S., Taigman, Y.: Make-a-video: Text-to-video generation without text-video data (2022)

\bibitem{song2022denoising}
Song, J., Meng, C., Ermon, S.: Denoising diffusion implicit models (2022)

\bibitem{vaswani2023attention}
Vaswani, A., Shazeer, N., Parmar, N., Uszkoreit, J., Jones, L., Gomez, A.N., Kaiser, L., Polosukhin, I.: Attention is all you need (2023)

\bibitem{wu2021godiva}
Wu, C., Huang, L., Zhang, Q., Li, B., Ji, L., Yang, F., Sapiro, G., Duan, N.: Godiva: Generating open-domain videos from natural descriptions (2021)

\bibitem{wu2021nuwa}
Wu, C., Liang, J., Ji, L., Yang, F., Fang, Y., Jiang, D., Duan, N.: N\"uwa: Visual synthesis pre-training for neural visual world creation (2021)

\bibitem{wu2023tuneavideo}
Wu, J.Z., Ge, Y., Wang, X., Lei, W., Gu, Y., Shi, Y., Hsu, W., Shan, Y., Qie, X., Shou, M.Z.: Tune-a-video: One-shot tuning of image diffusion models for text-to-video generation (2023)

\bibitem{yu2022vectorquantized}
Yu, J., Li, X., Koh, J.Y., Zhang, H., Pang, R., Qin, J., Ku, A., Xu, Y., Baldridge, J., Wu, Y.: Vector-quantized image modeling with improved vqgan (2022)

\bibitem{yu2022scaling}
Yu, J., Xu, Y., Koh, J.Y., Luong, T., Baid, G., Wang, Z., Vasudevan, V., Ku, A., Yang, Y., Ayan, B.K., Hutchinson, B., Han, W., Parekh, Z., Li, X., Zhang, H., Baldridge, J., Wu, Y.: Scaling autoregressive models for content-rich text-to-image generation (2022)

\bibitem{zhou2023magicvideo}
Zhou, D., Wang, W., Yan, H., Lv, W., Zhu, Y., Feng, J.: Magicvideo: Efficient video generation with latent diffusion models (2023)

\end{thebibliography}
\end{document}